# Breeding electric zebras in the fields of Medicine.[1]

Federico Cabitza[*,**], PhD

* Department of Informatics, University of Milano-Bicocca, Milan, Italy
**IRCCS Istituto Ortopedico Galeazzi, Milan, Italy

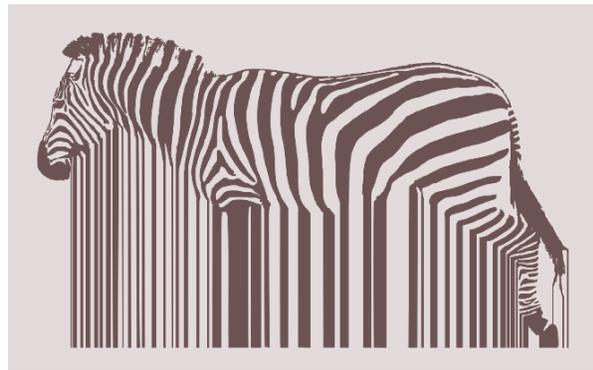

Figure 1. A zebra melting into a bar code. Artistic picture by Nevit Dilmen (2013, CC 0).

## A Tale of progress and opportunity

Rob is an excellent and expert radiologist. He is a physician who can use a wide set of advanced techniques in medical imaging like X-rays, CT, and MRI examinations to diagnose patients with many types of illness. Rob is in his late forties and many believe he will soon become the head of the radiology department of the hospital: when he makes a point in a meeting, few would contradict him. Many colleagues know that they can rely on him for an informed second opinion in case they have to cope with a difficult case. Some even believe that he is the most talented and knowledgeable radiologist of the hospital. For this job, Rob earns a quite good salary: last year this was $380,000, which yet is not extraordinary for radiologists with his expertise; according to the U.S. Bureau of Labor Statistics, that amount is just the median salary for physicians in his position. That notwithstanding Rob does not work to get rich. His passion for his job is genuine. To become such a radiologist, the competition for Rob had been long and fierce: after the high school, he had to get accepted for a 4-year university and then obtain a bachelor's degree with excellent medical school grades and very high university scores; then he had to collect a number of enthusiastic letters of recommendation on his attitude and competences to enter a medical school as a resident, where he practiced general medicine and surgery for one year before applying for a further 4-year training program in the field. During this latter program, Rob had to interpret, both day and night, tens of thousands of imaging studies, while also counseling patients and communicating and discussing results with his referring colleagues. Over time, his skills became clear and widely recognized at work.

Specialists like neurologists and orthopedists who refer patient to radiologists for digital imaging examinations usually disregard the radiologists' report when they consider the resulting images: they also know how to read images, do not want to be influenced in their decisions, and above all, do not want to reduce the analogical richness of an image to the discrete rigidity of verbal labels and ordinal categories reported in the radiological report. However, when the specialists of the hospital of Rob see that a report has been signed by him, they usually read it carefully and consider it trustfully.

Probably also for his reputation, one day Rob was involved in a reliability study in regard to the early diagnosis of Tuberculous Meningitis (TBM). This is a serious disease that usually kills two patients out of three and leaves one patient out of two with serious neurological consequences. In this study Rob, and other nine colleagues from several other hospitals, were supposed to assess a number of CT scans to assess and classify the entity of Basal Meningeal Enhancement (BME), that is a classic neuroradiologic features of TBM[2]. The results of this study were soon published and two particular aspects disquieted Rob. The intra-rater agreement (measured by means of the Fleiss' kappa statistic) assessed the extent Rob "agreed with himself" on the same CT scan being shown to him twice at different times during the day and in a random order with other control images: it was fair but not too high (0.55), i.e., he agreed with himself slightly more than 2 times out of three. However, the inter-rater agreement, that is the extent his findings were confirmed by the other colleagues involved in the study, was even lower (0.35), just half of the times. This result was not a thrashing of the radiologists' skills and knowledge. All the contrary, their diagnosis was still considered accurate as a whole, but showing some important variability: this could be traced back to a number of factors, like the reporting surgeon (and probably Rob was among the best raters), the device used to display the images, the radiologist's experience, and even the work shift (being agreement lower at the end of intense work shifts or by night). The authors of the study just commented on this point that the criteria for BME assessment needed to be standardized and validated in more thoroughly prospective cohort studies and that this kind of assessment was worthy of further study.

---

[2] This situation is inspired by the study reported in: Botha, H., Ackerman, C., Candy, S., Carr, J. A., Griffith-Richards, S., & Bateman, K. J. (2012). Reliability and diagnostic performance of CT imaging criteria in the diagnosis of tuberculous meningitis. PloS one, 7(6), e38982.

Some years later, the management of the hospital where Rob worked was contacted by an important IT company that proposed to provide the radiology department with a novel decision support system, called "Zebra Hunter"[3], a state-of-the-art machine learning system, in exchange of the expertise necessary to optimize it and the availability for hosting a series of experiments on its accuracy and reliability. However, the Zeb (how it was dearly called by its designers) was already very good in detecting anomalies and relevant traits of digital imaging: its accuracy for TBM cases, for instance, was 90% by means of a complex convolutional neural network. The management understood that this partnership could bring positive publicity and accepted to adopt this system in the idea that it could only improve the performance and accuracy of the radiological department. Moreover, it was decided that the Zeb should be used by radiologists only (and not, for instance by the neurologists): expert radiologists, like Rob, had to validate any of the Zeb's suggestions and decide whether to integrate them in their reports or not: these reports should not indicate whether the diagnosis was performed with the aid of the Zeb or without it. In other words, nothing should look different at the terminals of the department, although patients and physicians knew that the radiologists had an extra oomph for their daily job. Indeed, Zeb was used extensively and IT designers and radiologists worked intensively and closely together since its first deployment: its performance improved very soon, and its accuracy reached an impressive 98% on many types of images, while for others it was nevertheless above 85%: glowing figures that tied the best performances by the humans and that were totally irrespective of the work load, occasional resource shortages, the more or less frantic pace of hospital work. And mood swings.

The vignette can stop here. It already inspires us a number of questions. Some high-level questions are often addressed in both the Academic and white literature. One question is whether Zeb and systems like it will be used (and its bills be paid) more to increase hospital efficiency, prolong continuity of care, reduce health system costs, or improve care quality and outcomes. Moreover: are all these objectives positively correlated, or rather incompatible? Analysts also wonder if the almost sudden availability in the market of a number of "super Rob" (or at least, "tireless Rob") would cause a reduction of the salaries of expert radiologists like the real Rob, as well as of the demand by hospitals and hence the society for this medical role. As in the vignette above, we could think of the advent of a new medical role, probably extending the traditional skills of the radiologists, that can establish and maintain a trustworthy interface between the intelligent machine, kept as a sort of solitary oracle, and the rest of the socio-technical environment, that is anyone wanting to get access to the predictive services. This new radiologist would be an "information specialist"[4], i.e., someone who could properly feed in the machine and then interpret and make sense of its output and still filter and process it to make this "collaborative" advice be meaningful for the other physicians, the patients and all of the care givers. After all, approximately one century ago, this is exactly how the professional role of the radiologist was established and empowered: as an interface between the new X-ray equipment and the rest of the hospital physicians asking for the new type of consultation[5]. Moreover, exactly 40 years ago, Maxmen, an influential psychiatrist and professor at Columbia University, foresaw and advocated that within 2025 doctors would be substituted by a new "Medic-Machine symbiosis"[6] for the better provision of healthcare services.

However, we are more interested in subtler and more eerie questions: for instance, we wonder whether, and how, Zeb will affect the decisions of the doctors who use it, including Rob. Maybe Zeb would help novices learn how to interpret both easy and difficult images; improve the residents' skills more quickly also by challenging them with tricky simulations based on real cases; and teach also expert radiologists how to solve cases that before its arrival would be too difficult for a fast analysis and require several meetings, or worse yet, further examinations for the patient. Conversely, Zeb would perhaps undermine the self-confidence of the expert radiologists like Rob, after a few times that their interpretation differ from Zeb's and this latter proved to be the correct one. Likewise, it could make the young or less brilliant radiologists more lazy and dependent on its recommendations. Thus the main point here is whether systems like Zeb have the potential to actually deskill or "spoil" the physicians in the long run. After all, Rob and colleagues know that they can make mistakes, especially under time pressure or after 10 hours of work shift, while Zeb, also thanks to their daily feedback, has become almost 100% accurate on most kinds of imaging. Unfortunately, the radiologists know that Zeb will misclassify (either in terms of false negatives, or false positives) almost one case every day: they cannot let down their guard. Zeb is not infallible, but nevertheless its opinion must be held in very high esteem, just like Rob's opinion was, and still is, when he is at its best. More subtly, we also wonder if doctors will be allowed not to follow Zeb's advice. We are not speaking of a legal limitation, as accountability and responsibility will be at the human side for a long time still. Rather we are wondering whether ignoring Zeb will be socially or professionally blameworthy. Being against Zeb could seem a sign of obstinacy, arrogance, or presumption: after all that machine is right almost 98 times out of 100 for many pathologies, and no radiologist could seriously think to perform better than this. Probably novices and young radiologists would restrain themselves from defending their diagnostic hypothesis if different from the Zeb's one; but what about the seniors like Rob? Will using Zeb, and any similar system, either increase or decrease the number of "zebras"[7] that doctors will pursue whenever they hear hoofbeats? After all, Zeb will mention many alternative explanations all together with its best recommendation, even those that no radiologist would ever think of. Furthermore, how will this kind of decision support change the "political" status of radiologists like Rob towards the hospital management and other heads; the reputation-based hierarchical relationships among radiologists; the trust relationships between the radiologists and the other specialists; the tensions and collaborative dynamics between the IT staff and the physicians? Lastly, can we exclude that doctors will also use systems like Zeb to practice a stronger and more surreptitious defensive medicine, that is to choose for the most plausible option that defends them against potential controversies (whereas plausibility is estimated by the machine), or worse yet, to make Zeb a scapegoat to indulge in excusing both personal and teamwork failures (i.e., those related to collaboration and communication failures)?

---

[3] Unlikely as it may sound, the name for this fictional software is not based, nor even inspired, by the Imaging analytics platform by Zebra Medical Vision Inc. All Rights Reserved (www.zebra-med.com/). That notwithstanding, both systems ground on a famous medical saying, although probably from an opposite viewpoint.

[4] Jha, S., & Topol, E. J. (2016). Adapting to artificial intelligence: radiologists and pathologists as information specialists. JAMA, 316(22), 2353-2354.
[5] Reiser, S. J. (1981). Medicine and the Reign of Technology. Cambridge University Press.
[6] Maxmen, J. S. (1976). The post-physician era: Medicine in the twenty-first century.

We acknowledge and purport the open nature of these questions, as well as of any related question that any ethically engaged designer of medical technology should address. That notwithstanding, we will try to address some of them in the next sections.

**Explaining Machine Learning in the Ward**

It does not sound inappropriate to state that "data-intensive approaches to medicine based on predictive modeling"[8], what we will denote in what follows simply as Machine Learning (ML) has come to medicine to stay. However, the inquiring medical doctor could rightly ask, to go beyond –or better yet, behind - common buzzwords: what's machine learning? Put in very short terms, it's about learning by machines. The medical doctor could then urge: why the hell machines should learn? As it is clear to any computer science freshmen as well as experienced programmer, programming a machine, that is writing sets of instructions to have the machine do what we want, is a difficult, often frustrating and time-consuming task, which is also prone to many errors: thus, as noted by Mitchell[9], ever since the first computers were built, programmers wondered whether they might be made to learn by themselves, that is to improve automatically with experience. On the other hand, the idea that machines can learn is not that weird if one purposely sets aside the common notion that relates learning to knowing to adopt the viewpoint that relates learning to the improvement of performance. Research on machine learning then tells nothing about how people learn, but rather how machines can improve their performance over time. If this answer did not satisfy the medical doctor above (and she would not be entirely wrong), maybe further motivation could come from answering another related question: what can machines learn? In this case the short answer would be: the hidden structure binding data together (and probably so the related aspects of the reality that data represent). Discovering and making this structure explicit would be extremely difficult on the basis of formal logic, symbolic reasoning and mathematical modelling, which are all techniques that have been mastered by humans for centuries. On the contrary, a number of computational algorithms have recently proved to be very effective in detecting correlations and functional relationship among data, especially when these data are "big". The next question would follow naturally in the mind of the curious doctor: but how can a machine actually learn this hidden structure? Answering this question would entail some more explanations, as it regards the very idea of learning that is considered in the expression "machine learning" and how this idea and the related techniques can help medical doctors practice medicine and health care.

---

[7] Zebras in the medical jargon are the anomalies, the odd signs, the diseases so rare that most doctors never encounter them in regular medical practice. This term comes after the aphorism "When you hear hoofbeats, think of horses not zebras".

[8] Neff, G. (2013). Why big data won't cure us. Big data, 1(3), 117-123.

[9] Mitchell, T. M. (1997). Machine learning. 1997. Burr Ridge, IL: McGraw Hill, 45, 37.

Thus, put in general terms, any machine (and in fact any system, be it a software program or a human student), is said to learn in regard to a task T from experience E if, given a performance metrics P (also called objective function or scoring function), the latter's scores "improve" over time in carrying out T. A common example of P is counting the number of errors done over the number of attempts, whereas the system is said to learn from experience if P decreases over time. A Machine Learning (ML) approach just automatizes this process, which is called learning in this specific ambit, so that a machine, most of the times a software program, can play this game by itself, and improve its performance.

A sound ML approach regards the optimization of a decision model and doing this requires the definition of these three things: the task (to learn), T; the experience (to learn from), E; and how to evaluate the model, P: just a PET (remember this for later). Similarly, Domingos (2012) summarizes ML problems as the "combination of just three components": representation (basically of T and E); evaluation (i.e., computing P over a representation of T to see if it is good enough for some practical aim); and optimization (of the representation of T to achieve a better-scoring P). In traditional ML techniques applied to medicine, the task T is usually medical classification (usually applied to prognosis, but also to diagnosis and treatment choice increasingly), and the experience E is a set of cases already classified by human experts. From a computational perspective yet, the task T is usually tantamount to the production (or selection) of one value, y, (be this either a nominal class or a number) from a set of possible values (Y), given an x from an n-dimensional domain X. In other words, T can be represented as a functional mapping from X, the data we possess, to Y, the data we want to predict or infer from X, i.e., a function f so that $y = f(x)$. This mathematical way to see the completion of a task should not look odd to a layman, neither to a medical doctor. This is is just a general way to see both classification and regression tasks, for instance, the diagnosis of a medical case or the prediction of a lab exam result. Computer scientists like to use a functional representation whenever a task can be formulated in terms of the association of either a category or a number (the 'y') with a set of data (the coordinates of a point 'x', that describe or represent a case on a multidimensional space of attributes or features). ML techniques aim to approximate this function f by inducting hypothetical functions (a set of f', or the hypothesis space) from a limited set of mappings ($y_i=f(x_i)$) by improving the score of an evaluation function P applied to a f'. The set of mappings is what we above called experience E (also training experience), that is the "knowledge" that it is available in regard to the task at hand. ML algorithms (learners) are said to learn the underlying structure binding together X and Y, that is the "real" function f. Now, computer scientists usually are concerned with the increasing and efficient optimization of the classifiers/predictors, that is the f', modelling the above structure.

**The Unintended Consequences of Machine Learning in Medicine**

So far so good. In the third and last part of this work-in-progress we want to focus on the potential unintended consequences of using ML in support of medical practice. However, we will not consider the consequences of mistakes by ML algorithms like those that recently hit the headlines for patent discrimination based on either race[10] or gender[11]. Rather, we want to focus on the unintended consequences of ML at its best, when its accuracy in automatic classification is considered at least acceptable or even comparable to (if not even higher than) that of human practitioners. This kind of exercise is not completely new: A comprehensive framework of Unintended Consequences (UC) was developed by Ash almost ten years ago[12] in regard to health IT and then also applied to computerized order entry[13]. We believe that many

---

[10] Barr, A. (2015) Google Mistakenly Tags Black People as 'Gorillas,' Showing Limits of Algorithms. The Wall Street Journal. http://archive.is/EE8PM. Accessed on 01/11/2017

[11] Datta, A., Tschantz, M. C., & Datta, A. (2015). Automated experiments on ad privacy settings. Proceedings on Privacy Enhancing Technologies, 2015(1), 92-112.

[12] Ash, J. S., Berg, M., & Coiera, E. (2004). Some unintended consequences of information technology in health care: the nature of patient care information system-related errors. Journal of the American Medical Informatics Association, 11(2), 104-112.

[13] Campbell, E. M., Sittig, D. F., Ash, J. S., Guappone, K. P., & Dykstra, R. H. (2006). Types of unintended consequences related to computerized provider order entry. Journal of the American Medical Informatics Association, 13(5), 547-556.
Ash, J. S., Sittig, D. F., Poon, E. G., Guappone, K., Campbell, E., & Dykstra, R. H. (2007). The extent and importance of unintended consequences related to computerized provider order entry. Journal of the American Medical Informatics Association, 14(4), 415-423.Ash, Joan S., Dean F. Sittig, Richard of the types of unintended consequences considered in that framework could also be related to accurate (and we emphasize accurate) ML-driven decision support systems (MLDSS), especially: "untoward changes in communication patterns and practices; negative emotions; generation of new kinds of errors; unexpected changes in the power structure; and overdependence on the technology". That notwithstanding, here we want to introduce some potential ailments that are more peculiar of the extensive use of MLDSS, like: *empirical anopsia* and *conversational hypoacousia,* which both regard the representation of the available Experience (E) and the input of the MLDSS; *probabilistic tinnitus*, which regards the representation of the outcome of the task (T) and hence the output of the MLDSS; *epistemic sclerosis*, which regards the representation of the performance of the task itself; *metric paroxysm*, that regards the very way performance is assessed (that is, P); and lastly, *semiotic desensitization*, *judgemental atrophy* and *oracular rush* which derive from relying too much on the "accurate" output of the MLDSS and its effect on the interpretative work of medical doctors. Many of these problems are related to a common cause: overreliance. This in its turn comes from an excess of trust in the power of ML models and the DSS based on these latter models. This trust is stoked, on one hand, by the biases that Greenhalgh[14] dubbed "pro-innovation" and "human substitution": that is believing that a new technology is inherently better than anything already in use and that, once a task has been automated, the automating technology must be as good as, or even better than, any human involved in that task. On the other hand, this trust is influenced by any kind of literature, both white and academic, reporting very high accuracy rates for the latest ML implementations in healthcare, ranging from 80% to 99% [RIF]. This has spread a sort of *metric paroxysm*, that is an ill-grounded trust in numeric measures of the performance of ML-based classifiers. While the most wide spread metrics of such a performance, that is accuracy (i.e., error rate), is probably also the most deceptive one (cf. the accuracy paradox[15]), the specialist literature is still looking for the best concise way to express a classifier quality and compare classifiers: specificity, sensitivity, precision, the F-score (that is the harmonic mean of sensitivity and precision), the Youden index, the diagnostic OR/DOR, the Area Under the ROC curve (AUC), just to mention the most common ones. Still too many ways to shoot at a moving target. However, as noted by Kuhn and Johnson[16] the value of a ML model emerges more from a complex trade-off between its accuracy and its explainability, which must be evaluated qualitatively, rather than by means of a single figure. But even more radically than this, the quality of a medical DSS should be probably traced back to the improvement of outcome and, to guarantee this on the long run, satisfaction and ease of the decision makers (i.e, the medical doctors). Paradoxically less accurate MLDSS could be better, in terms of impact on outcome, than the mostly accurate ones, e.g., because the former ones do not undermine the doctors' self-confidence or disrupt consolidated team practices (more on this later). Value is to be found in the ratio between benefits (for patients, doctors, and hospital managers, either tangible and intangible ones) and costs, the economic, the cognitive and the failure-related ones. In this light, even simple and common measures of technical accuracy, like sensitivity and specificity, must be taken with caution and possibly seen beyond their mere numeric value to be considered for what they actually mean in situated and embodied clinical practice: e.g., by considering the extent false negative rates reassure human diagnosticians about their indispensability; false positive rates makes them increasingly mistrustful (and hence willing to quit using the system or disuses it, even if it brings value); and low error rates make them increasingly *reliant* on the allegedly accurate aid. In particular, Parasuraman and Riley[17] observed how overreliance on (any) technology can lead to either its misuse or abuse. Abuse is use irrespective of the stunning lack of evidence base for the efficacy and cost effectiveness of DSS in the Dickensian world[18] where medical doctors work on a daily basis, that is the repression of the clues that should instead make people wary of the actual capability of technology to really improve outcome, a subtype of overreliance that could be called

---

Dykstra, Emily Campbell, and Kenneth Guappone. "The unintended consequences of computerized provider order entry: findings from a mixed methods exploration." International journal of medical informatics 78 (2009): S69-S76.

Harrison, M. I., Koppel, R., & Bar-Lev, S. (2007). Unintended consequences of information technologies in health care—an interactive sociotechnical analysis. Journal of the American medical informatics Association, 14(5), 542-549.

[14] Greenhalgh, T. (2013). Five biases of new technologies. Br J Gen Pract, 63(613), 425-425.

[15] Valverde-Albacete, F. J., & Peláez-Moreno, C. (2014). 100% classification accuracy considered harmful: The normalized information transfer factor explains the accuracy paradox. PloS one, 9(1), e84217.

[16] Kuhn, M., & Johnson, K. (2013). Applied predictive modeling (pp. 389-400). New York: Springer.

[17] Parasuraman, R., & Riley, V. (1997). Humans and automation: Use, misuse, disuse, abuse. Human Factors: The Journal of the Human Factors and Ergonomics Society, 39(2), 230-253.

[18] Coiera, E. Evidence-based health informatics. https://coiera.com/2016/02/11/evidence-based-health-informatics/ Accessed on the 06/01/2017.

overconfidence. However, since prediction for either the better or the worse is always difficult, (especially about the future, as ironically noted by Niels Bohr), a more blameworthy attitude seems another type of overreliance, i.e., overdependance, or misuse. This is the use of the technology beyond actual needs, just because it's either more handy, faster or just more convenient. In so doing, users can end up by forgetting any contingency plan, that is any alternative system that could substitute the automated one whenever this fails, is interrupted or breaks down. Even worse, misuse can weaken the skills (or deskill[19]) of misusers that are related to the automated function in the long run. In particular, we call *judgemental atrophy* the progressive process of deskilling of medical doctors in regard to the ability to arrive to a sound conclusion and knowledgeable opinion by following a process of observation, reflection and analysis of observable signs and available data. We refer to judgement, because this ailment does not regard only, nor primarily, a cognitive impairment regarding knowledge and medical competence, but more subtly a lack of self-confidence, an enervation of the will to take a stance and also, sometimes, even some risk. Thus, this UC regards the side effect of any interference of MLDSS on the very content or timing of the process of decision making, be this either individual or collective in case of medical teams, about a clinical case. It is difficult to envision how this class of new DSS will affect the physicians' self-confidence, hierarchical power relationships or the relationship with patients. That notwithstanding, we can inspect this broad phenomenon in some more details by distinguishing between *empirical anopsia, conversational hypoacousia* and *probabilistic tinnitus*: the former two regard the input, the latter one the output of the models yielded by ML computing. In particular, empirical anopsia and conversational hypoacousia are sensory defects of the doctors that come up by "overemphasizing structured and complete"[20] data entry and information retrieval with respect to a more holistic assessment of the patient's conditions. Let's see each of these potential shortcomings in some more details. *Empirical anopsia*[21] is a consequence of overrelying on the available data as a trustful representations of the clinical phenomena the data relate to. While it is undisputed that any sensor, including the human eye, is sensitive to only a limited amount of the perceivable stimuli, and rightly so depending on what it is considered relevant to take a certain course of action, empirical anopsia occurs when there is a lack of awareness of the essential arbitrariness of the aspects that any data structure and its values depict of a real and continuous phenomenon, as well as of those elements of this phenomenon that any of its representations amplifies, conceals or, worse yet (necessarily), distorts. It is the partial "blindness" of considering not only the available data but any possible data as a sufficient "proxy" and substitute of the phenomenon that these are supposed to render into a discrete (often either numerical or codified), transportable (mobile) and immutable form[22].

If *empirical anopsia* regards the intrinsic and -to some extent- unavoidable inadequacy of the representation of medical cases (broadly speaking), that is of the knowledge base that constitutes the "training experience" (E) on which machine learning algorithms rely on to build progressively accurate classifying and predictive models, *probabilistic tinnitus* is a more specific condition. It regards the output of these classifiers and predictors. As widely known, tinnitus is the perception of a ringing noise in the ears that affects the perception of external sounds. This noise can be either low pitched or high pitched: in the former case we can speak of low-frequency tinnitus, while in the latter case we can speak of this disorder as acting as a sort of high-pass filter that attenuates or eliminates the perception of sounds with frequencies lower than the cutoff frequency. Speaking out of metaphors, this condition represents the potential harmful consequences of two typical ways to present ML results, that is either as clear-cut categories or sharp numerical values (e.g., normal/pathologic, diagnosis A/diagnosis B), or with a ranking of possible resulting classes or values expressed in probabilistic terms (e.g., diagnosis A 78%, diagnosis B 10%). Likewise, *probabilistic tinnitus* is a disease coming in two variants: when results are displayed in the ranking form there is *low-frequency tinnitus* in that low-probability conditions, i.e., the sound of improbable but yet possible the zebras, can affect the perception of the hoofbeats of the more plausible horses, and hence prejudice the interpretation of medical doctors in also considering what, without the DSS, they would have never thought of (thus leading to possible overprescription of unnecessary diagnostic examinations); on the other hand, when results are presented in terms of clear-cut categories, *high-pass probabilistic tinnitus* can occur, as higher probability diagnoses simply end up by hiding a whole spectrum of less probable but yet possible options, thus leading to a form of over-simplification of complex cases that could nevertheless present some important comorbidity. *Conversational hypoacousia* also regards the representation of the case but, as the reference to the hearing hints at, this ailment relates closely to the patient-doctor relationship: it regards the fact that doctors using an ML DSS could be even less interested (than commonly assumed [RIF]) in the discursive element of the symptom reporting by the patient, its proxemic dimension, as well as the narrative one, as all of these elements are the less suitable ones to be fed into the DSS and usually not as much useful to enable accurate ML-based processing as structured data, codes, and numeric measures instead are. This problem closely relates to the following one.

---

[19] Carr, N. (2015). The glass cage: Where automation is taking us. Random House.

[20] Ash et al. (2004). Op. Cit.

[21] Anopsia is a medical term that indicates any defect in the visual field, from enlarged blind spot to colour vision deficiencies.

[22] Latour, B. (1986). Visualization and cognition. Knowledge and society, 6(1), 1-40.

*Semiotic desensitization*. With this expression we call the progressive decrease of responsiveness and sensitivity of physicians with respect to the material, bodily, analog and continuous *signs* of the patient who stands in front of them, in favor of the discrete data proposed by electronic patient records, registries and decision support tools. These data are the measured and digitized (that is, often numerical but also categorical) counterparts of the patient's signs, which are produced by any kind of automated probe, sensor and device. Therefore, semiotic desensitization is a consequence of the "quantified patient"[23]. This tantalizing representation of the patient, seen as a set of features, is just the "big" amount of data that it is necessary to make machine learning tools increasingly accurate and reliable; that is to make systems like the DSS the optimal *intermediary* between the actual body of the patient who turns to doctors for help and care, and the required decisions that doctors have to make on how to intervene on that body to solve the patient's problems. Oracular rush regards the attitude by which doctors could tend to consult the machine as soon as they can, after collecting few elements of the anamnesis and of the physical examination, and sooner than a more conservative approach, known as "wait and see" would suggest to wait for clearer signs and confirmed diagnosis. This could lead to overdiagnosis and overtesting, which it is known bring negative consequences on the patient wellbeing [RIF].

Lastly, but perhaps the most subtly dangerous consequence: *Epistemic sclerosis*[24] . This is a specific risk related to the application of machine learning to human classification: as said above, these are techniques of supervised classification that "learn" the underlying structure that binds the above mentioned empirical data (i.e., features) to their categorical interpretation (predetermined classes). This operation is aimed at representing what we can know about these things, their functional relation and apply this (optimized) *knowledge* to new observations. Epistemic sclerosis occurs whenever this simple operation risks to freeze the always-to-some-extent arbitrary, unreliable, and idiosyncratic mapping between the sign-data dyad and the *right class* identified by one (or few) observers. These latter observers, as widely known but often overlooked in the Academic literature, many times just do not agree with each other, and sometimes even with themselves: this is the known case of both inter- and intra-rater reliability, or observer variability. Often in the medical literature variability is not considered a problem at all, as long as clinicians achieve an agreement and solve opinion conflicts. However, as pointed out by[25], to decide what the "true" category is on a majority basis cannot be appropriate if raters are few; moreover, doing it on a consensus basis is not entirely free of risks, both in case consensus is achieved after open discussion or by anonymous voting. Researchers should consider if observer variability is not related to an intrinsic ambiguity of the phenomena observed, rather than to an interpretative deficiency. However, the impact of observer variability on ML cannot be overestimated. Indeed, a ML approach risk to freeze into the decision model two serious and often neglected biases: *selection bias*, occurring when training data (the above experience E) are not fully representative of the natural case variety due to sampling and sample size; and *classification bias*, occurring when the single categories associated by the raters to the training data oversimplify borderline cases (i.e., cases for which the observers do not agree, or could not reach an agreement), or when the raters misclassify the cases (for any reason). In both cases, the decision model would surreptitiously embed biases, mistakes and discriminations and, worse yet, even reiterate and reinforce them on the new cases processed. If ML specialists were more aware of this, they could become more wary of classifications where the original data present relevant inter- or intra-rater disagreements. In those cases, MLDSS designers could explore the opportunity to transform dichotomous values into a multi-class range of four values: a 1/0 domain (e.g., pathological vs. normal) could be transformed into one including the following values: 1, 0, 1-and-0 and even a "*mu*"[26]

---

[23] Smith, G. J., & Vonthethoff, B. (2016). Health by numbers? Exploring the practice and experience of datafied health. Health Sociology Review, 1-16.

[24] In pathology as well as in botany, a sclerosis is a hardening of a tissue, usually by thickening or lignification.

[25] Svensson, C. M., Hübler, R., & Figge, M. T. (2015). Automated Classification of Circulating Tumor Cells and the Impact of Interobsever Variability on Classifier Training and Performance. Journal of immunology research, 2015.

[26] "Mu" may be used similarly to "N/A" or "not applicable," following the popular interpretation by Robert Pirsig and Douglas Hofstadter of the

value. The former case could be used when an even number of raters could choose one option instead of the other; the latter could be used whenever clinicians recognized cases that are really borderline and that could not be settled on the basis of a majority-based consensus. What ML specialists often ignore is that many times physicians reach diagnosis on a "best guess" basis and likely according to elements that either are not represented properly in the available data nor represented at all. Thus the mapping between training experience (E) is fraught with an uncertainty that current ML algorithms just do not address. Thus, accepting that reality can be difficult if not impossible to be classified can be key when a black-box model of ML risks to freeze a forcedly orderly association between data and their "standard" interpretation and make doctors more careful in their judicious decisions. In regard to the *black-boxness* of ML models, this is often considered *the* problem of the use of ML in DS. It is argued that MLDSS should not be evaluated only in terms of accuracy (i.e., error rates), but also in terms of interpretability, in primis by the doctor enacting the decision. This latter dimension is much harder to assess, and various synonyms have been proposed: explainability, intelligibility, transparency, understandability[27]. Ironically enough, the more the models take biological-inspired names (e.g., random forests, neural networks), the less understandable the mapping[28], not only for doctors and laymen, in general, but also for the engineers themselves. The point is not that MLDSS do not give explanations for their suggestions: after all these explanations could be either selected from a predefined set or composed ad-hoc by a further automatic mapping. The problem rather lies in the impossibility for the doctor to "open" the black box and uderstand *why* certain classes should be associated to certain patient features, let alone "try" himself (like in a sort of simulation) the mapping between a case and the most appropriate decision instead of having to rely on blind faith. Paradoxically, what in a human context could be said to relate to creativity, ingenuity and intuitivity[29], in a MLDSS context could be seen as a problem of opacity and, worse yet, ill-grounded trust, i.e., trust on the capabilities of an actual "Clever Hans horse"[30], rather than on a reliable decision companion. Overfitting models (that is models very well trained on a sample of all possible data that yet do not properly represent the whole population) could be "horses" that are much better in getting their carrot than helping their master do math.

**Conclusions**

Obviously the risky consequences mentioned above should not be taken as disjoint categories and indeed we see them as strongly correlated and mutually affecting, leading to known phenomena like overdiagnosis, overtesting and overprescription, but also distreatment (i.e., avoiding risky treatments for a sort of defensive medicine) and obviously misdiagnosing (mistaking the diagnosis before contrary evidence). We conjecture that many times their coexistence can let even paradoxical phenomena emerge. For instance very accurate MLDSSs could, in the long term, deskill those who are supposed to maintain, and update them. Deskilling could negatively affect their ability to interpret and classify new images (cf. semiotic desensitization) that regular maintenance would require to feed into the systems as training data to increase their accuracy. The paradox lies in the fact that these systems that undermine the skills of their janitors could become less and less accurate over time, although fewer and fewer people could recognize this by that time[31].

Those risks can be considered typical of a new generation of accurate decision support systems, as they are related to their very essence: DSS are systems designed to help doctors take decisions, that is make up their mind by "cutting off" (quite literally) the unlikely and inappropriate options, so that the *best ones* remain and can be selected and executed. These systems are then aids of optimal *option selection*, among the alternatives that have been predefined in some theoretical model, according to more or less explicit data-driven conditions. Here suffice to hint at the idea that adopting a "choice support system", rather than a DSS, would entail a completely different mindset: whereas choice denotes "the right, power, or opportunity to choose", such a system would be aimed at helping physicians have a perception of what the right or wrong choice may be, pondering the odds and risks, being free to choose what, to the best of her knowledge and beliefs, she deems better for

---

famous Zen Buddhism koan whenever categorical thinking is considered to be an unnecessary stretch or even a delusion. To this respect we also recall the famous Tao Te Ching saying: "The five colors [that is thinking of colors in terms of only five linguistic categories] make man's eyes blind" (cf. Lecture on Zen by Alan Watts)

[27] Lipton, Z. C. (2016). The mythos of model interpretability. arXiv preprint arXiv:1606.03490.

[28] Carstensen, J. (2016) Is Artificial Intelligence Permanently Inscrutable? Despite new biology-like tools, some insist interpretation is impossible. nautil.us/issue/40. http://nautil.us/issue/40/learning/is-artificial-intelligence-permanently-inscrutable. Accessed 01/26/2017. Archived at http://archive.is/geMja.

[29] after all doctors are not supposed, or maybe just should not always be supposed, to say why they feel that a certain decision is more appropriate.

[30] B. L. Sturm, "A simple method to determine if a music information retrieval system is a 'horse'," IEEE Trans. Multimedia 16(6):1636–1644, 2014.

[31] personal communication with Mireille Hildebrandt, 2016.

the patient at hand, according to the latter's needs and wishes.

As noted by Neff (2013, cit.), the biggest challenge for the use of ML in health care is social, not technical. The challenge is still to understand how people, both patients and providers, will actually embed automated classifications and prediction into practice. In this light, to consider the potential consequences of machine learning like fanciful diseases is not just a way to dispel with irony the odds of their occurrence. More than that, it is a metaphoric way to suggest that, like with biological disease, we can produce the necessary antibodies to fight these potential shortcomings related to the application of new predictive technologies to medicine, and that prevention, based on better knowledge and awareness, is far better than cure. Thus, far from being a Cassandra's exercise, or worse yet, of any neo-luddite, the research on the unintended consequences of any technology adoption finds its most natural justification on the raising of awareness by a larger portion of its prospective users and a better sensitivity on the connected nature of sociotechnical settings: that is the awareness that if you change something, even a small thing in a process, unexpected and uncontrolled changes can spread to elements that one could not suspect to be connected together: this seems particularly true especially in those settings where technology can act as a *third actor* coming between doctors and patients[32], thus affecting the quality and function of this relationships between humans. We believe that the prudent attitude of a UC research can reduce the odds of the negative consequences. If these latter ones, right in virtue of their unexpectedness, occur all our efforts notwithstanding, UC research can help us manage them and reduce their impact.

In addition to a prevident attitude, UC research can also inspire a different approach to health IT design. Inspired by this approach, we aim to evaluate the effectiveness of some tentative (technological) antidote. In particular, we are investigating if visualization tools[33] that provide doctors with alternative, visual, number-less (i.e., analogical) representations of the patient's conditions and perceptions, illness evolution and treatment risks (see Figure 2) could support their choices and also get them more wary of the validity of clear-cut categories and quantitative numbers. In so doing, we also aim to reinforce the idea that medical practice is primarily still an "art and science of the signs".

Exactly two hundred years ago, Lander-Beauvais[34] conceived and proposed to his colleagues this vision of what medicine should be. Now we are wondering if it is important to preserve such a vision, in this age in which machines learn how to *datafy* medical signs better and faster than any brilliant student in a Medical School, actually can learn to interpret them.

**Acknowledgements**

The author is deeply grateful to Raffaele Rasoini, MD cardiologist, and Prof. Giuseppe Banfi, scientific director of IRCCS Galeazzi, for the fruitful talks and conversations on the relationship between medicine and ML technology.

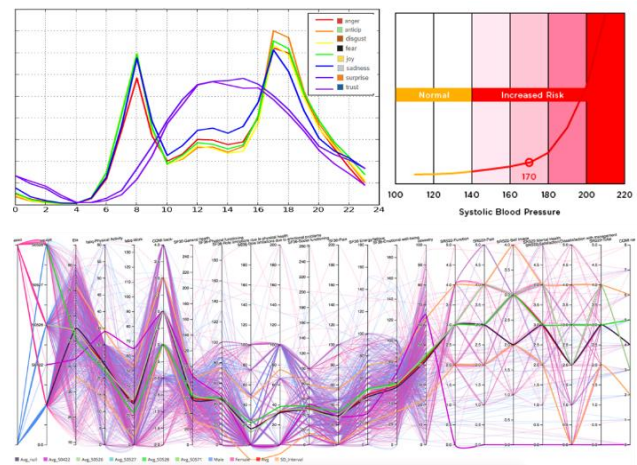

Figure 2: Examples of choice-aiding visualizations: (from top left, in clockwise direction) a *rhesiogram*, that is a graphical representation of how basic emotions (like anger, fear, trust, sadness and joy) unfold in the patient's writings; a picture to render the risk associated with blood pressure (from www.vizhealth.org); a parallel coordinate diagram showing multivariate patient attributes.

---

[32] Swinglehurst, D., Roberts, C., & Greenhalgh, T. (2011). Opening up the 'black box' of the electronic patient record: A linguistic ethnographic study in general practice. Communication & medicine, 8(1), 3.

[33] Cabitza, F., Locoro, A., Fogli, D., & Giacomin, M. (2016). Valuable Visualization of Healthcare Information: From the Quantified Self Data to Conversations. In Proceedings of the International Working Conference on Advanced Visual Interfaces (pp. 376-380). ACM.

[34] Landré-Beauvais AJ (1818). Séméiotique, ou traité des signes des maladies. Paris: J.A. Brosson